# Summative Student Course Review Tool Based on Machine Learning Sentiment Analysis to Enhance Life Science Feedback Efficacy

Ben Hoar,*[1] Roshini Ramachandran,†[1] Marc Levis,†[2] Erin Sparck,†[3] Ke Wu,*[2] Chong Liu*[3]

*UCLA Chemistry and Biochemistry {bhoar@chem.ucla.edu[1], coco.wuke.sz@gmail.com[2], chongliu@chem.ucla.edu[3]***}

†UCLA Center for the Advancement of Teaching {rramachandran@teaching.ucla.edu[1], mlevis@teaching.ucla.edu[2], esparck@teaching.ucla.edu[3]}

*** please direct communications to chongliu@chem.ucla.edu

*Machine learning enables the development of new, supplemental, and empowering tools that can either expand existing technologies or invent new ones. In education, space exists for a tool that supports generic student course review formats to organize and recapitulate students' views on the pedagogical practices to which they are exposed. Often, student opinions are gathered with a general comment section that solicits their feelings towards their courses without polling specifics about course contents. Herein, we show a novel approach to summarizing and organizing students' opinions via analyzing their sentiment towards a course as a function of the language/vocabulary used to convey their opinions about a class and its contents. This analysis is derived from their responses to a general comment section encountered at the end of post-course review surveys. This analysis, accomplished with Python, LaTeX, and Google's Natural Language API, allows for the conversion of unstructured text data into both general and topic-specific sub-reports that convey students' views in a unique, novel way.*

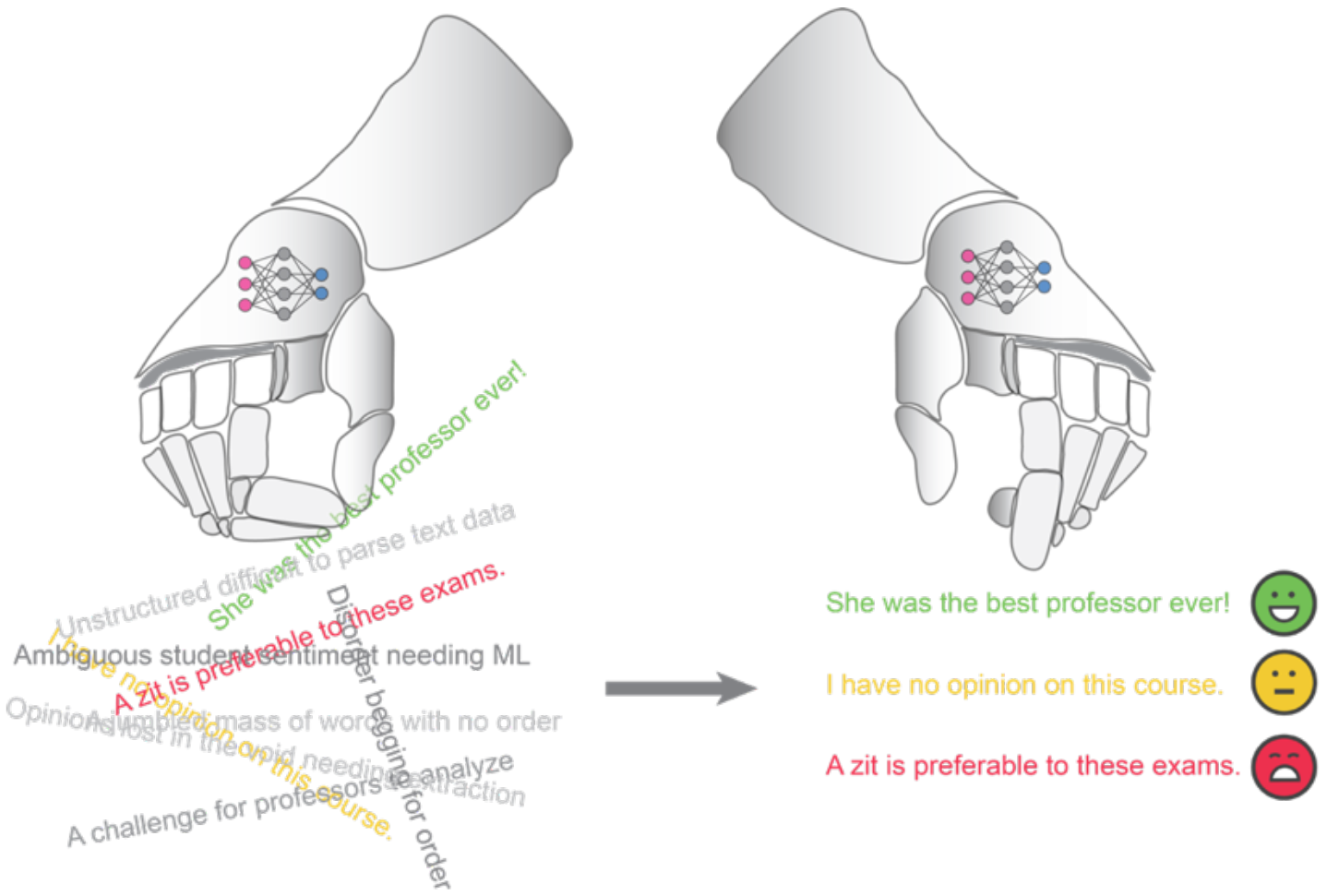

## *Introduction*

Student course reviews are a fundamental way in which students participate in their own education.[1, 2] By providing feedback at the end of their tenure in a course, students allow professors to process what they are doing well and what they may improve on, hopefully improving course quality for both educators and students over time. One limitation of this approach, however, is that students are often asked to provide feedback in an unstructured, open-ended manner.[3, 4] This approach is challenging in one key way – it does not organize the data into an easily digestible format that can be quickly analyzed by professors. When data is unstructured, patterns in student sentiment may be missed, ignored, or falsely derived from biases in the mind of the reader.[5, 6] Because of this, a tool to organize and present student reviews in a way that provides both general and topic-specific insight would be highly useful. One such way to accomplish this goal is via the combination of text-processing/programming principles combined with state-of-the-art machine learning techniques. Text-processing and programming can provide a means to split unstructured data into subcategories and machine learning techniques can be utilized to score data on a "sentiment scale" that ranks the split data (e.g. phrases and sentences) on a numerical one to five scale that correlates to a gradient of negative to positive sentiment.[7, 8] In the following text, discussion will focus on how Python programming principles are combined with the Google Cloud Sentiment Analysis API (GCSA), as well as how text data was gathered and used to train our machine learning model (Figure 1). Finally, professor feedback will be presented to gauge this tool's efficacy in its current state.

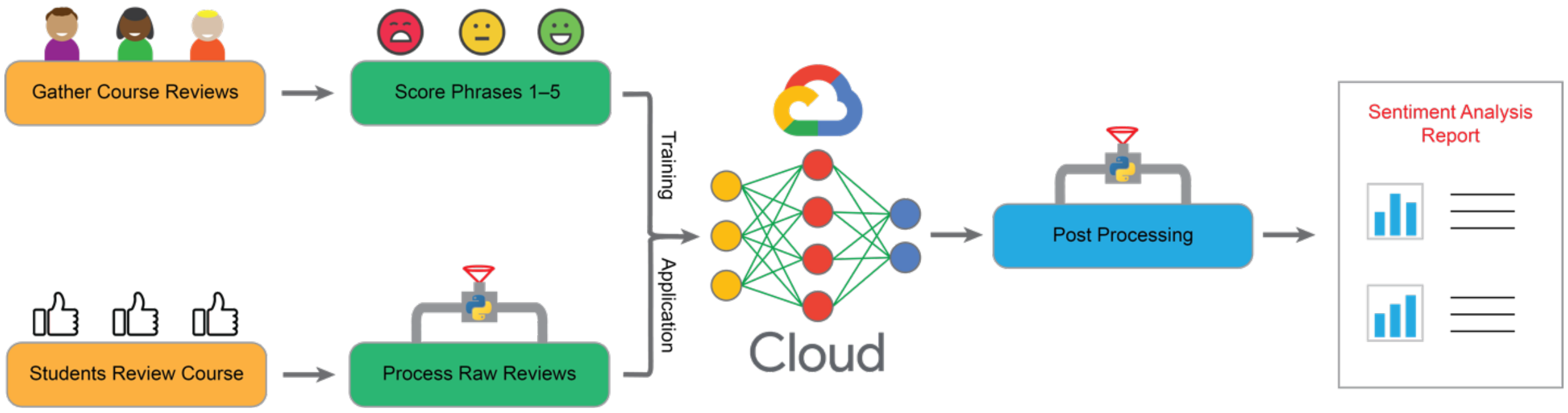

**Figure 1:** Overview of training and general application pipeline. In the algorithm training phase, course reviews are gathered, scored, and used to train a GCSA algorithm which can then be subsequently used to generate a sentiment analysis report. After training, student course reviews can then be processed and scored only by the GCSA algorithm and then used for report generation.

## *Background*

Prior to the discussion of the specifics of this tool, it is important to discuss what machine learning and sentiment analysis are to better understand the value that this technique provides. Generally, machine learning is defined as "the use and development of computer systems that are able to learn and adapt without following explicit instructions, by using algorithms and statistical models to analyze and draw inferences from patterns in data."[9] This definition implies that if there are patterns in data that correlate to an output of interest, machine learning can be used to train algorithms to find this correlation and subsequently be used to analyze future data. A subset of machine learning, sentiment analysis, aims to learn how the vocabulary and language employed by an author can be used to score their opinions as positive, negative, or somewhere in the middle.[7] This technique is ubiquitous in companies aiming to gauge public opinion on their products or marketing strategies.[10] Often, these opinions are gathered from public platforms such as Twitter and Instagram and subsequently analyzed using sentiment analysis algorithms. Once opinions are gathered and scored, easy to interpret graphs and tables can be used to succinctly demonstrate how well a company's product, advertisements, etc. are being received by the public. One well known case within this context is the analysis of IMDb movie reviews, a source of highly opinionated text data that is ubiquitous in the testing of sentiment analysis algorithms and in student learning of sentiment analysis techniques.[11-13] Further, sentiment analysis is not limited to corporate product monitoring as it has also aided with the analysis of politicians' communications about the COVID-19 pandemic and how their tone impacts the efficacy of messaging and information spread.[14] Given the success of sentiment analysis techniques in these contexts, it seems logical that they could be applied to student course reviews as courses are essentially a product to be reviewed or perhaps marketing for academic subjects.

To implement these text sentiment techniques, various paths can be taken. These options include training and engineering a machine learning algorithm from scratch, using a pre-trained sentiment analysis algorithm, and using a service such as GCSA. In this paper, the latter of those two options are employed as they are the most flexible and "easy" to implement for researchers wishing to employ these techniques with only a minor background in machine learning – making this technique highly accessible to those with programming capabilities.

## *Results*

With the accessibility of both pre-trained algorithms and GCSA in mind, the key step to implementing these techniques (this discussion primarily focused on the GCSA), is to gather and score data representative of cases that have arisen and will arise in the future.[15] Machine learning functions on learning the correlations between data and some known output so that future data can be input into the algorithm and have its output predicted without human interference, greatly accelerating[16, 17] the analysis of large quantities of data. In the case of this research, the input is student reviews in the form of statements or sentences and the output is the relative positivity of these statements on a scale of 1–5. To gather this data, various UCLA professors and teaching assistants were solicited to provide student reviews of their courses from previous quarters. Due to the generally positive nature of these reviews, *ratemyprofessor.com* was also consulted to extract more negative and neutral reviews to balance our training data. In the end, this resulted in a dataset of 1603 phrases (Figure S1). To minimize bias in the scoring of the dataset (initial data collection and approximate scoring was performed by Benjamin Hoar), a total of 20 UCLA undergraduate students were gathered to perform additional scoring. Directions on scoring stated that scoring should be conducted as objectively as possible[18] – meaning that the language should be used as the basis for scoring and not personal biases, as best as possible. For example, students often comment that a course is "fast-paced," which is *not inherently* negative or positive, it is merely a reflection of the course structure. On the other hand, a statement claiming the course to be "too fast-paced and therefore difficult to follow" is a direct criticism of the course. In the end, the median of the total of 20 scores was counted as the *true* value of the phrase (Supplementary Methods). After scoring, this data was split into sets containing 1,282 training, 161 validation, and 160 testing instances. This split in data is necessary to select the best algorithm structure, with the training data used to extract correlations in the data and output, the validation set used to test the accuracy of the model *during* training, and the testing set used to test the accuracy of the model *after* training.[10] Using this split, the GCSA tool provided a trained sentiment analysis algorithm with an overall accuracy of 73.1% (Figure S2), in line with state-of-the-art[8, 19] accuracies for fine-grained sentiment analysis (fine-grained meaning non-binary scoring).

Following training, the sentiment analysis algorithm could be called from within custom python scripts developed specifically for student course review supplementation. These python scripts provided the means to convert any given raw text source into the format required for input into the scoring algorithm and subsequently its organization into a summative report (Supplementary Report) for professor review. The first stage of this report generation involves the splitting of raw text data into phrases. This is accomplished by splitting on standard sentence-ending punctuation and the word "but." Following this, each statement could be converted to a standard form and then scored by the GCSA algorithm and saved for subsequent report generation.

The two types of output for report generation are the "general" results section and a "topic-specific" results section. The "general" results section provides an overview of the entire text corpus and reflects student sentiment as a function of all student review statements. Here, four graphs are presented (Figure 2). The first two of the four graphs show a summary of student sentiment both as a function of student sentiment normalized *by author* (Figure 2A) and a second showing *all* statements' sentiment regardless of author (Figure 2B). The key purpose for the former of those graphs is to remove bias resulting from a highly verbose author (e.g., an outlier student writing many negative reviews would skew the 2B graph, but not 2A). Succeeding these histograms are two more summative graphs. Students in UCLA life science courses are asked to rate the course on a scale of 0-9, giving an overall final score for their experience in their respective courses. In Figure 2C, the distribution of these scores is plotted versus the same distribution as presented in 2A, giving two ways of viewing student sentiment – one derived from a simple numerical scale and one derived from their written opinions. Finally, figure 2D encompasses the "topic-specific" results section. In this "topic-specific" section, pages show student sentiment as it pertains to course aspects such as the lecture, homework, professor, etc. However, before this section, the results presented therein are summarized in a way to give an overview of how each key course aspect was received.

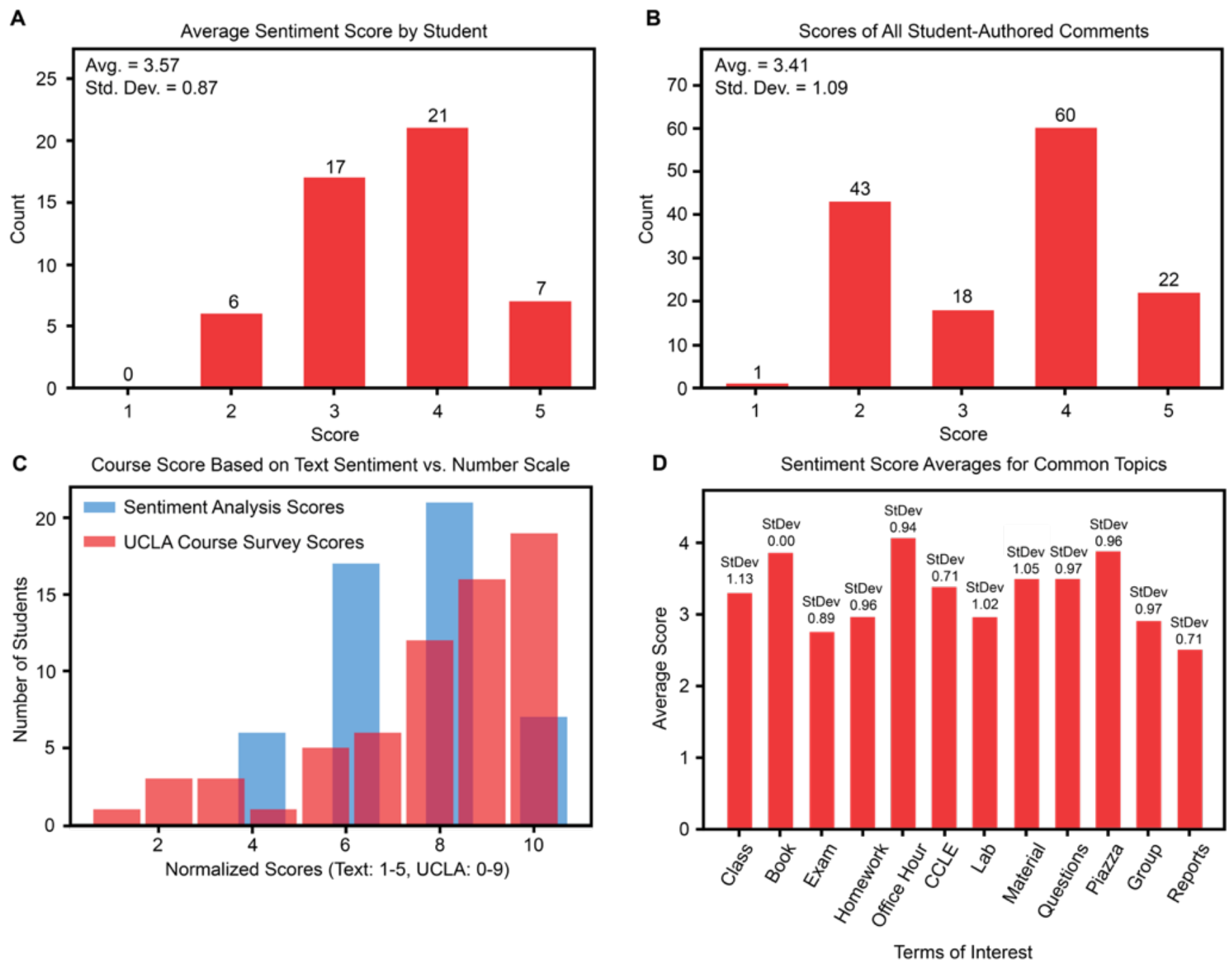

**Figure 2:** Panel A shows average student sentiment of each student in the course, while panel B provides insight into the raw count of each score in the entire text corpus. Panel C aids in visualizing the difference in raw numerical scores given by students when prompted to rate the course versus their implied score as a function of their language. Panel D summarizes sentiment regarding course aspects.

In addition to these summative reports, "topic-specific" reports were also generated to provide insight into how specific course practices were received. Two methods of generating these reports were considered. First, via a discussion between UCLA professors and advisors from the UCLA Center for the Advancement of Teaching, a list of globally relevant terms that encompassed course aspects that are common to most, if not all, university level life-science courses was established. The terms covered the instructor, the lecture, the textbook, the exams, homework, CCLE (an announcement, grade, and document student portal), and office hours. In addition to these terms, six auto-selected terms were also used to generate topic-specific reports. These terms were selected via a word-cloud inspired algorithm (Figure 3A).[20] In a word cloud, the most common words in a body of text are presented in a graphic, with the most common words appearing larger. To select for these terms, the counts of all words of the entire student-review text corpus were calculated. From this list of word-counts, words already accounted for in the aforementioned topic reports and meaningless "stop-words"[21] (words such as the, if, and, etc.) were also removed from this text body. What is left over is a list of the most referenced terms. Using these terms, reports were generated for topics most relevant to the specific course as dictated by the topics of greatest interest to the students according to their reviews.

Considering these terms of interest, each topic is given a page to demonstrate the overall sentiment of the students as it pertains to that topic, as well as giving a subset of actual phrases from the students. On each of these pages a graph providing a distribution (i.e., Figure 3B) is presented atop a table (Figure 3C) of exemplary phrases – providing samples from each of the five scores of the ranking system. The phrases selected for presentation were further fed through a binary sentiment analysis algorithm, VADER (Valence Aware Dictionary and sEntiment Reasoner),[22] that provides secondary assurance that the presented scores were accurate representations of their classes. The desire for this is to mitigate the effect of errors stemming from the initial algorithm and provide quality control for the phrases presented in these "topic-specific" reports. Once phrases were selected, they offer the professor insight into overall sentiment about a topic (Figure 3B) and specific statements that indicate what students may have liked and not liked about that topic's execution (Figure 3C). Once all the data processing, scoring, and organization was accomplished, the data could be automatically converted into a standalone report with the aid of pyLaTeX,[23] a python library that can be implemented to create LaTeX documents automatically (Supplementary Report). Within this document, directions on how to interpret the data and additional insight into its generation were also included.

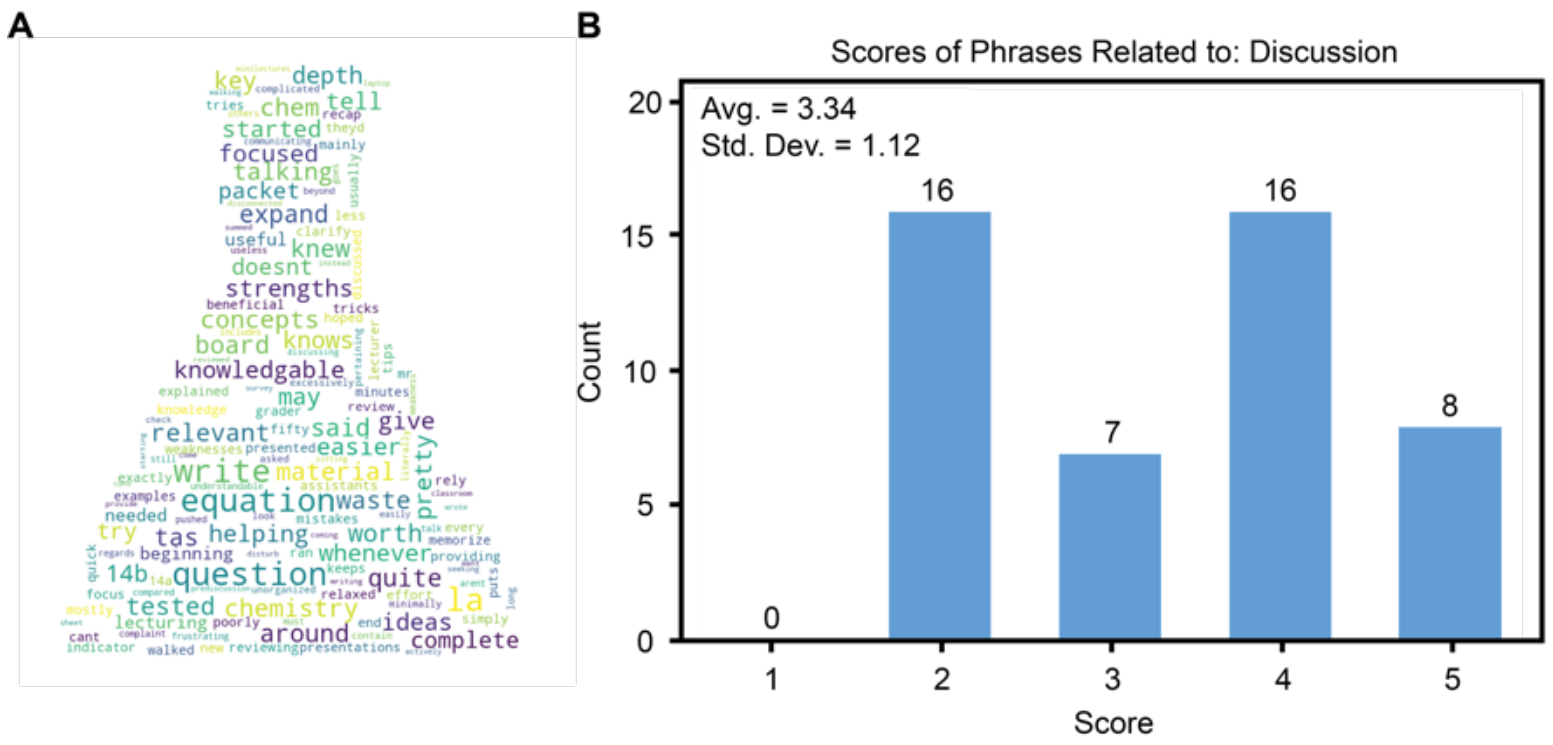

C

| Comment | Score |
|---|---|
| He was an outstanding TA always willing to understand our situations and expand on course material. | 5 |
| He taught so efficiently and covered a lot of ground in a single discussion. | 4 |
| It seemed like we only went to discussion to work on a worksheet on our own. | 3 |
| Sometimes he would be a bit slow, worksheets would have helped. | 2 |
| He made a poor impression during discussion and his input was unhelpful and smug during lectures. | 1 |

**Figure 3:** Panel A visualizes the coding principle used to automatically select for terms presented as relevant topic-specific sub reports. These reports each contain a summative graph as shown in panel B and a table of comments in panel C that constitute the comments on this course aspect.

To gauge the utility of this proof-of-concept supplementary report, four UCLA professors were polled about their opinions on this tool. In general (Table 1), the report was found to be useful, but certain limitations and concerns exist with the current format that may be remediated with future iterations of this method. One significant update to this tool – one that could address issues surrounding both the topics picked and how representative the tabulated comments are of those topics – is to emancipate it from the standalone pdf format (as with the Supplementary Report). In the background, all student phrases are scored, but there may be too many to make a concise report, especially when considering the addition of six topic-specific reports. A solution to this is to migrate this format to a web-based tool that allows professors to look at overview statistics, but also to have autonomy over which terms are used for the topic-specific reports (terms could be suggested to the user, but also requested by the user). This would allow a professor to, for example, obtain the reviews and histogram for a topic in their class that may have been missed in the current format. For instance, a professor may be concerned with the grading curve in their class. In a following iteration of this tool, the professor could poll the scored data for comments on the curve simply by inputting the term “curve” into some search bar. This would offer professors the ability to see both student sentiment on common course aspects that are suggested by the tool, but also opinions on more niche class aspects they may be interested in.

Furthermore, any issues concerning the accuracy of scoring may become less worrisome in this free-form format as a web-based tool could present *all* phrases associated with a class aspect and not just a sample that is currently used to allow for the automated generation of LaTeX documents. With a sample, it is possible that a few mis-scored statements make it into the table as representatives of a score and consequently lower user confidence in the tool’s ability. While there is an inherent challenge in performing fine-grained sentiment analysis, the algorithm does a strong job of predicting negative comments as negative and positive as positive. Because of this, a tool that presents all statements of a topic rather than a subset that may randomly contain an unsatisfactory number of phrases scored in error would prove less likely to create confusion or alarm in the mind of the reader. Another remedy to any concerns with the accuracy of the scoring of the algorithm could be to gather more training data, but this is an expensive process in terms of human-hours.

| | Strongly Disagree | Disagree | Agree | Strongly Agree |
|---|---|---|---|---|
| The report's overall organization and structure were satisfactory. | 0 | 0 | 4 | 0 |
| The introductory comments and descriptions helped me to understand how the report was generated. | 0 | 1 | 3 | 0 |
| The graphs were clear to interpret. | 0 | 2 | 2 | 0 |
| Organization by keyword was useful. | 0 | 0 | 4 | 0 |
| The keywords selected reflected the most important components of my course. | 0 | 2 | 1 | 1 |
| The algorithm used to generate the reports adequately categorized the responses (i.e. the number assigned to each comment generally matched the sentiment). | 0 | 2 | 2 | 0 |
| I am satisfied with the 5-point scale. | 0 | 0 | 4 | 0 |
| The comments told the entire story of what students felt happened in my class (based on existing report format). | 0 | 2 | 1 | 1 |
| This report was a useful complement to the existing course review format. | 0 | 0 | 3 | 1 |
| The representative comments provided in the report were sufficient to help me improve my course. | 0 | 2 | 1 | 1 |

**Table 1:** Summary of four professors' opinions on this tools' efficacy.

## *Conclusion*

In short, a proof-of-concept tool has been developed to aid professors in their efforts to better understand and adapt to the sentiments of their students. By combining software-engineering and machine learning principles, professors can now visualize how the language used by students in their course reviews relates to their feelings about their course in a quantitative way. Furthermore, in current course review practices, students only comment on what they are prompted on outside of the "general comments" section, providing a challenge for professors to analyze how well specific aspects of their course are being received. With this tool, professors can analyze via graphical and tabular information the opinions of their students with respect to the most common course aspects of interest. In the future, this tool can be expanded via emancipation from the pdf output format and moving it to a web-based tool to create a greater ability for professors to get a custom feedback experience.

## *Supplementary Methods*

Sample size determination is critical in any empirical study in which the goal is to make inferences about population parameters from sample estimators.[1] In this study, the goal is to select an appropriate sample size which accurately represents the target population, the total number of students enrolled in life science courses at UCLA within an academic year. Given a 95% confidence level, a 5% margin of error, and an approximate population size of 1500 students, the minimum sample size to estimate the true population proportion with the required margin of error and confidence level is 306 students. The calculation of sample size (n) used the following formula: $n = \frac{N*X}{(X+N-1)}$, where $X = \frac{Z_{\propto/2}{}^{2}*p*(1-p)}{E^{2}}$. "n" is the population size (=1500), $Z_{\propto/2}$ is the critical value (=1.96) of the normal distribution at $\propto = 0.05$, p is the estimated sample proportion (=0.5), E is the margin of error (=0.05). However, affected by the factors of cost, time, and convenience of data collection, we randomly selected 20 student participants from life science courses, and their scoring of the database containing 1603 phrases served as the preliminary data used to train sentiment analysis model in GCSA.

The discrepancy between the calculated sample size (n=306) and the used sample size (n=20) can be counterbalanced by counting the median of the total 20 scores as the true value for each teaching evaluation phrase. If the lower bound of the confidence interval of the proportion of 20 participants scoring the same number is above 50%, we are 95% confident that the median of the total 20 scores can truly represent the median of the entire population. To calculate the minimum sample proportion (p) required to satisfy this condition, the formula $n = \frac{Z_{\propto/2}{}^{2}*p*(1-p)}{E^{2}}$ was used, where n is the sample size (=20), $Z_{\propto/2}$ is the critical value (=1.96) of the normal distribution at $\propto = 0.05$, E is the margin of error (unknown). The result showed that when the proportion of 20 scores being the same number is above 70%, the resulting margin of error is 20% and thus the 95% confidence interval must be larger or equal to 50%, making the median score of our selected sample size valid and unbiased. By this metric, 40% of our training data scores can be declared statistically accurate; however, in practice, the scoring requirements were not rigid enough for this issue to be disqualifying for a proof-of-concept tool. This metric is valuable to track, though, for as the tool's use evolves, gathering of greater numbers of student labelers may help increase both the accuracy of the trained tool and the confidence with which to tool is used.

## *Supplementary Figures*

**A**

| Score | Count |
|---|---|
| 1 | 138 |
| 2 | 433 |
| 3 | 279 |
| 4 | 447 |
| 5 | 306 |

**B**

| Dataset | Count |
|---|---|
| Training | 1282 |
| Validation | 161 |
| Testing | 160 |

**Figure S1:** Class populations of the various sentiment analysis scores (A) and sizes of the various datasets used in the training and testing of the model (B).

| Actual Score \ Predicted Score | Sentiment Score 1 | Sentiment Score 2 | Sentiment Score 3 | Sentiment Score 4 | Sentiment Score 5 |
| --- | --- | --- | --- | --- | --- |
| Sentiment Score 1 | 50% | 36% | 14% | - | - |
| Sentiment Score 2 | 2% | 79% | 9% | 7% | 2% |
| Sentiment Score 3 | - | 18% | 68% | 11% | 4% |
| Sentiment Score 4 | - | 4% | 4% | 76% | 16% |
| Sentiment Score 5 | - | - | - | 23% | 77% |

**Figure S2:** Confusion matrix of testing data after sentiment analysis model training. Percentages along the diagonal are correct classifications and other cells provide insight into types of errors made by the model.

## *Supplementary References*

## *Supplementary Report*

# Student Evaluations Sentiment Analysis

Reviews for: [redacted]

March 31, 2021

## 1 Overview

The following is the sentiment analysis report of your student reviews from:

19S_CHEM_14B_DIS_[redacted].csv 20W_CHEM_14B_DIS [redacted].csv
20W_CHEM_14B_DIS_[redacted].csv 18F_CHEM_14A_DIS [redacted].csv

Scores were calculated using a predictive model trained via the Google Natural Language Sentiment Analysis API. The model was trained on real student course reviews sourced from UCLA faculty and RateMyProfessor.com. The data used for training was scored on a scale of 1 to 5 (negative to positive) by twenty-one UCLA students, the median score of the students was subsequently taken as the "real" score for each review. In this report, your raw review data has been processed, scored, and organized into summative and categorical sections. This review processing is similar to the example shown below:

**Original Review:**
*The professor was very knowledgable. I thought she/he was very capable, but often they did not provide enough direction. Overall, though, I enjoyed the class!*
**Parsed Reviews:**
*The professor was very knowledgable.*
*I thought she/he was very capable*
*Often they did not provide enough direction.*
*Overall though, I enjoyed the class!*

These parsed reviews, along with their scores, can be filtered and used to generate this report – aiming to enhance the existing review process by highlighting student sentiment towards various course aspects. Scores presented herein are predicted by the Google Natural Language API Trained model. The model has a 73% accuracy, however errored predictions typically fall within one point of the correct score (~92% of instances are either correct or within 1 point).

1: More Negative – 2: Negative – 3: Neutral – 4: Positive – 5: More Positive

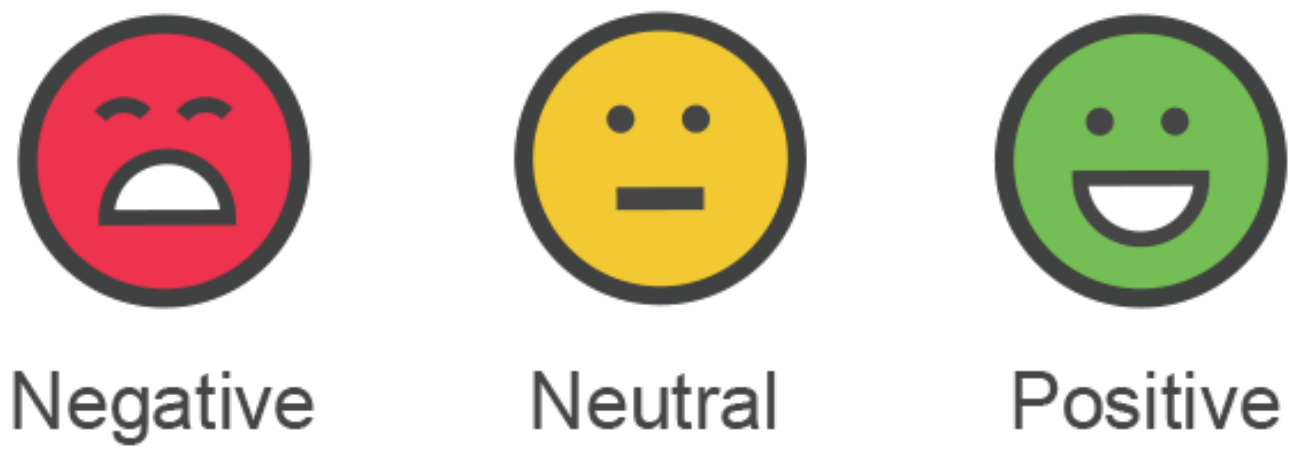

## 2 Scoring System

The following is a more detailed discussion of the scoring scheme. Please, not that the comments presented on this page are NOT from your course evaluations and are intended only to help introduce the the scoring system. As mentioned on page 1, the scale was initially determined by human scorers in order to train the algorithm that subsequently provides the scores presented in this and other reports. Please, note that the algorithm makes mistakes, but provides general guidance and insight into student opinion on course aspects – globally and specifically – providing strong aggregate results based on algorithm training results. Errors tend to be more pronounceed for the most negative reviews (false 1 predictions) when the true sentiment is biased highly positive. For example, over the course of a full body of reviews, some scores of 1 are bound to be given to statements as a function of misclassification. Due to the algorithms behind this report, those misclassifications are especially likely to be shown if the comments are overall heavily biased towards the positive end. We wanted to show examples of each, but if no students said anything negative, then the comments scored in error will appear as the representatives of those scores. A brief explanation and visual of the scoring system is given below. Phrases presented in the visual were scored by the algorithm:

**Score 5:** These scores are considered to be exceptionally laudatory or positive.

**Score 4:** These scores are considered to be generally positive or favorable.

**Score 3:** These scores are considered to be neutral, or highly *subjective or ambiguous* in terms of favorability. For example, a comment claiming a course was "very fast-paced" may be considered a criticism to some, or neutral to positive to others.

**Score 2:** These scores are considered to be generally negative or critical.

**Score 1:** These scores are considered to be more unfavorable or negative.

| | Sample Reviews – De-Identified | Score |
|---|---|---|
| More Positive | I believe that she is honestly the best chemistry TA out there. | 5 |
| | The instructor made sure to ask and help others and was very easy to talk to in regards to the course material. | 4 |
| Approx. Neutral | The tests were only focused on being able to complete math/ chemistry related problems. | 3 |
| | He did not seem like he wanted to be there and sometimes sounded that he was just explaining the material to himself. | 2 |
| More Negative | He made a poor impression during section and his input was unhelpful and smug during lectures when asked questions. | 1 |

# 3 General Results

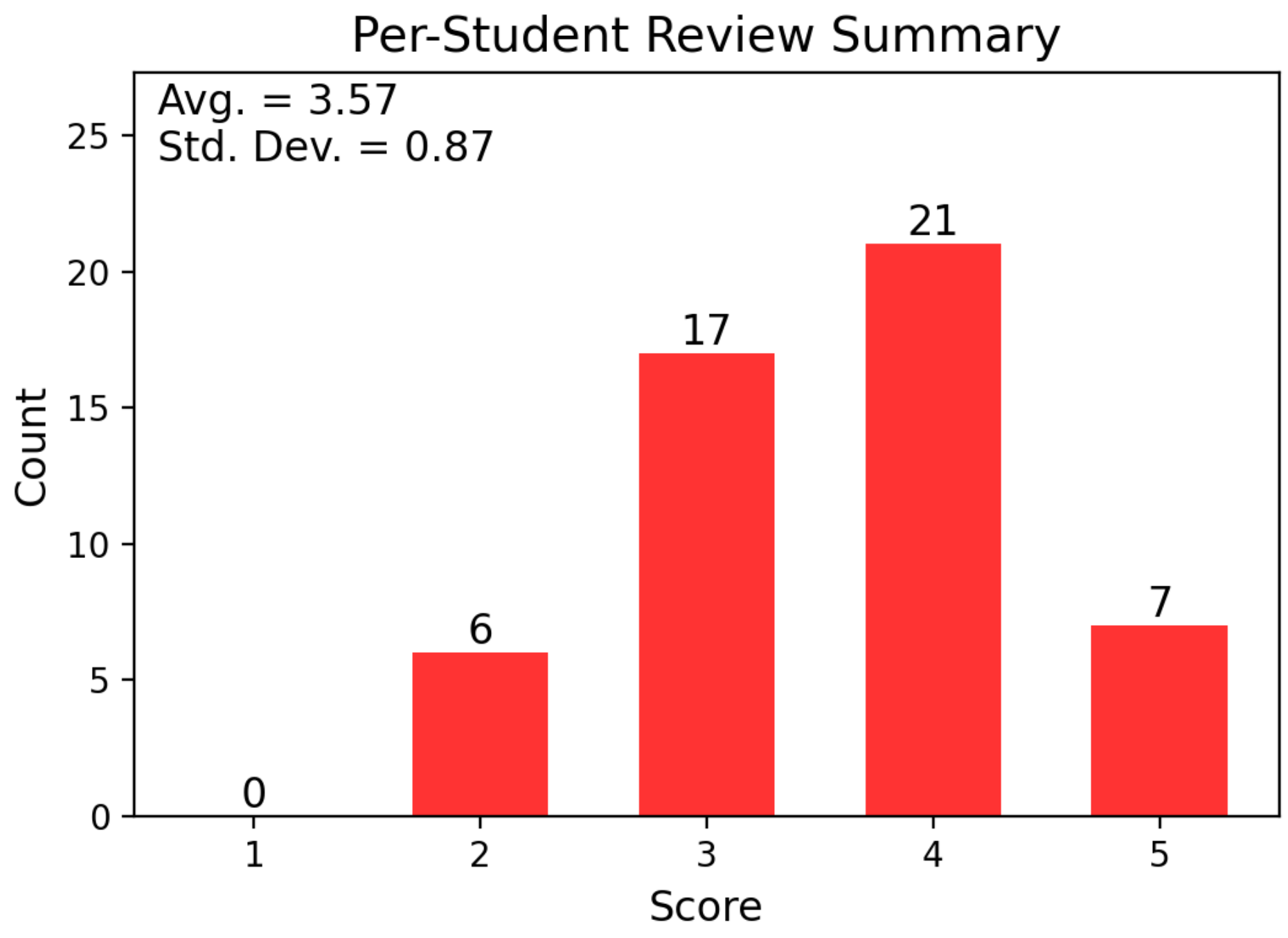

Figure 1: Scores of phrases averaged by their student authors. For example, one student may have made five comments with an average score of 4. That average is represented here.

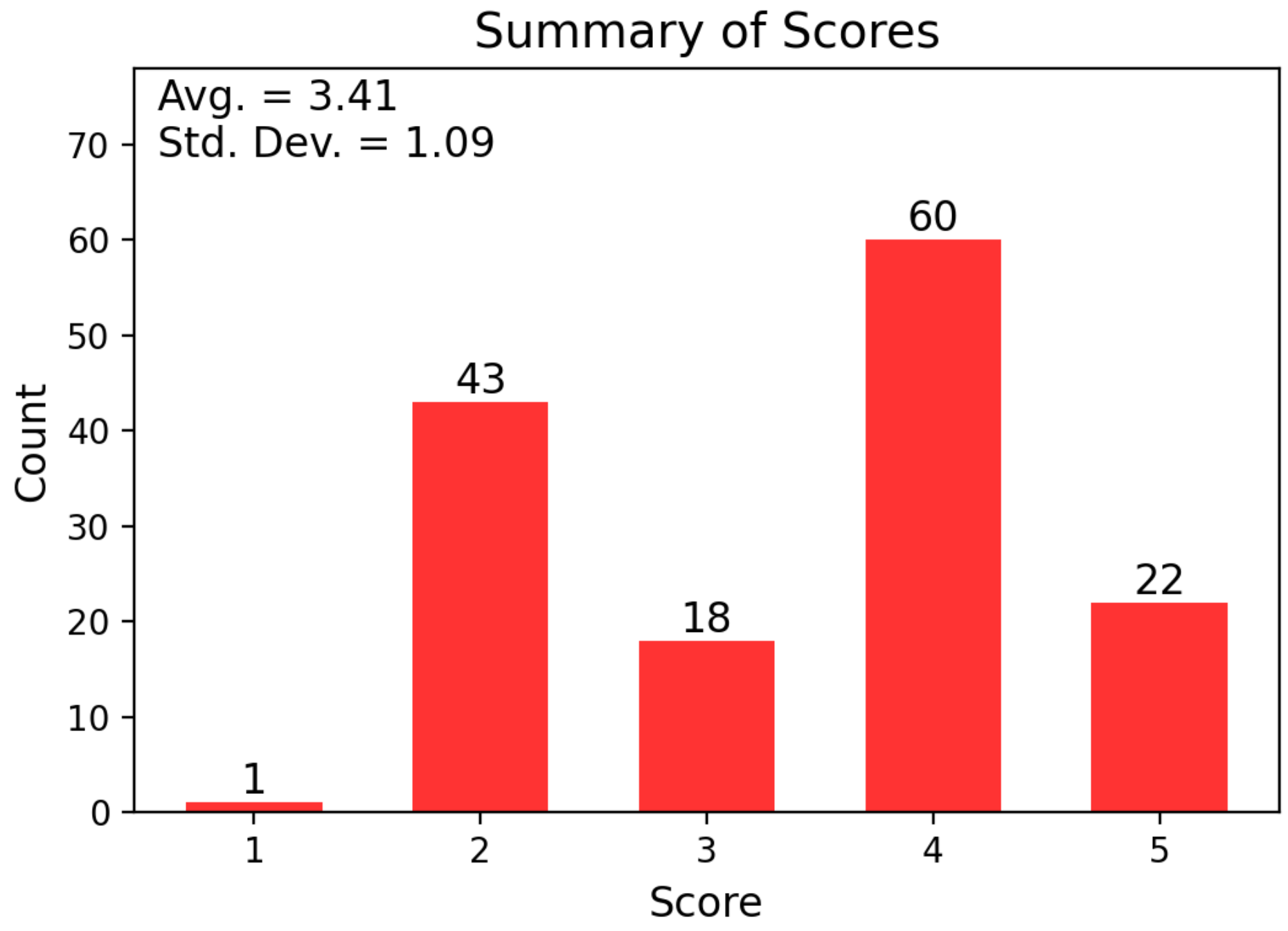

Figure 2: Raw scores of all phrases from all students. These are the scores for all parsed comments from students regardless of author. For example, if all students made 25 comments that were scored as 5, then the x-axis value of 5 would show a count of 25.

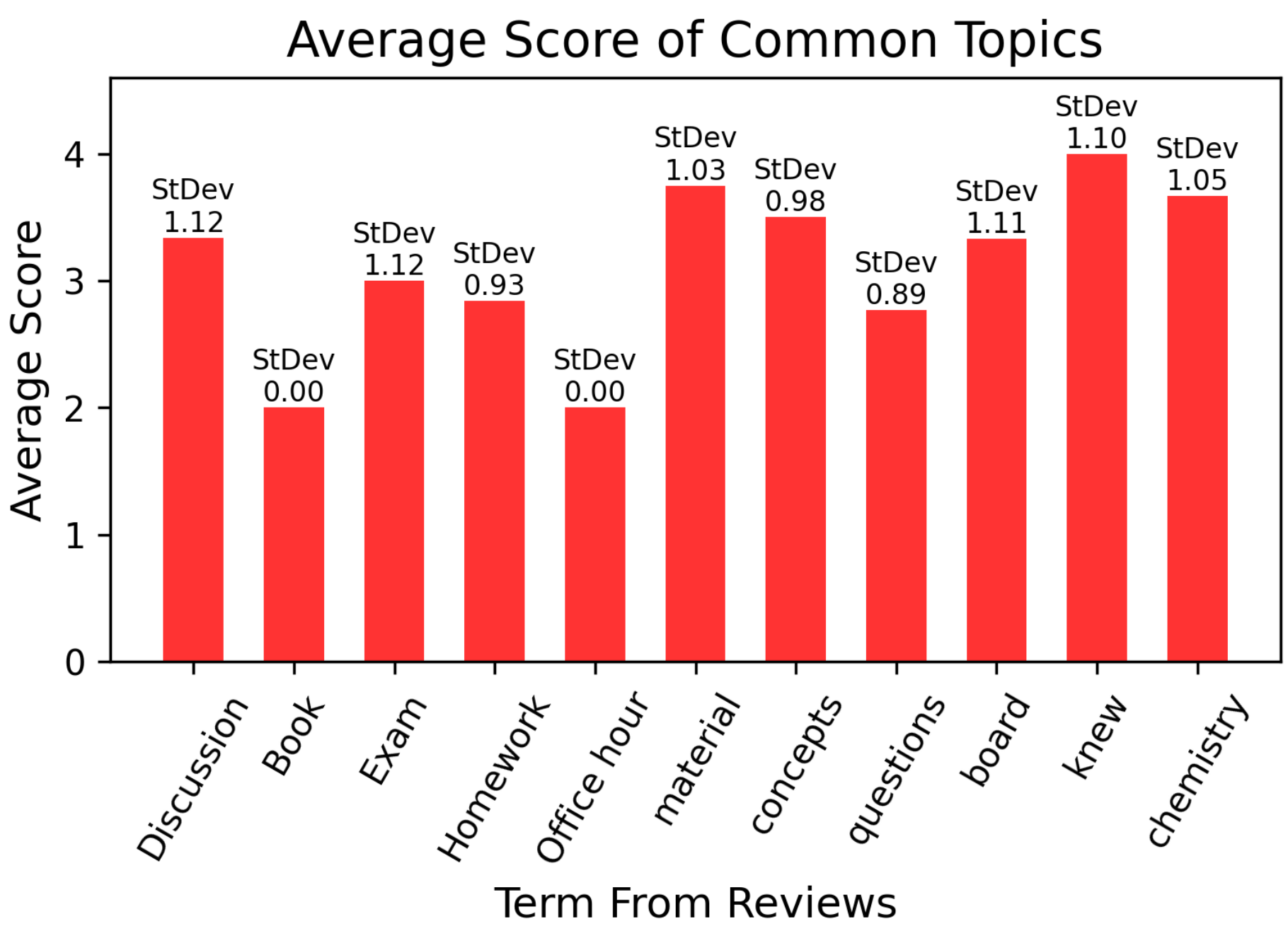

Figure 3: Average scores of phrases referencing common topics. The first 6 terms on the x-axis were manually chosen for their relevance to all courses, and the rest were chosen by an algorithm discussed on the next page.

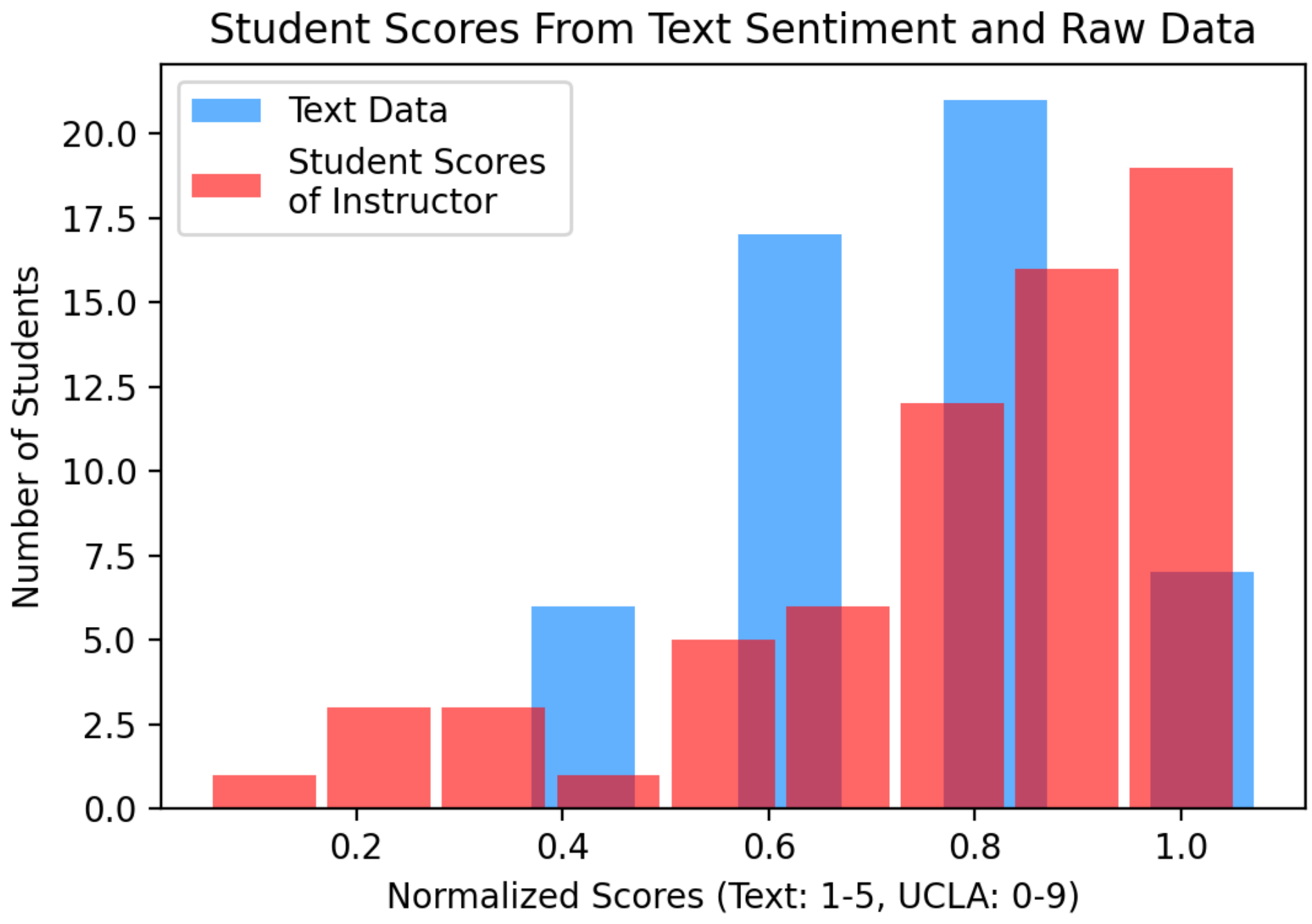

Figure 4: Overall course evaluation score distributions based on sentiment analysis (blue, see Figure 1) versus values provided by students in their course evaluations (red). Note – both scales were adjusted to a 0-1 scale for comparison purposes.

# 4 Word Specific Summaries

The following pages, until the end of the report, are word specific reports. These terms were selected in two ways as shown in the subsections below.

## 4.1 Pre-Selected Terms

The terms shown here represent aspects of the course that are considered of general interest; these terms were pre-selected by committee. Words that are grouped together are considered synonymous.

[[redacted], 'They', ' He ', 'She ', 'instructor', 'lecturer', 'teaching assistant']
['Discussion', 'class', 'course']
['Book']
['Exam', 'midterm', 'final']
['Homework', 'hw ', 'worksheet', 'practice problem', 'assignment']
['Office hour', ' OH ']

## 4.2 Auto-Selected Terms

The figure shown here is a word-cloud of common terms (besides the ones already listed above) and was used as a foundation for selecting terms not encompassed by the above lists. The top 6 terms are presented in "word-reports" below.

The auto-selected terms are: ['material', 'concepts', 'questions', 'board', 'knew', 'chemistry']

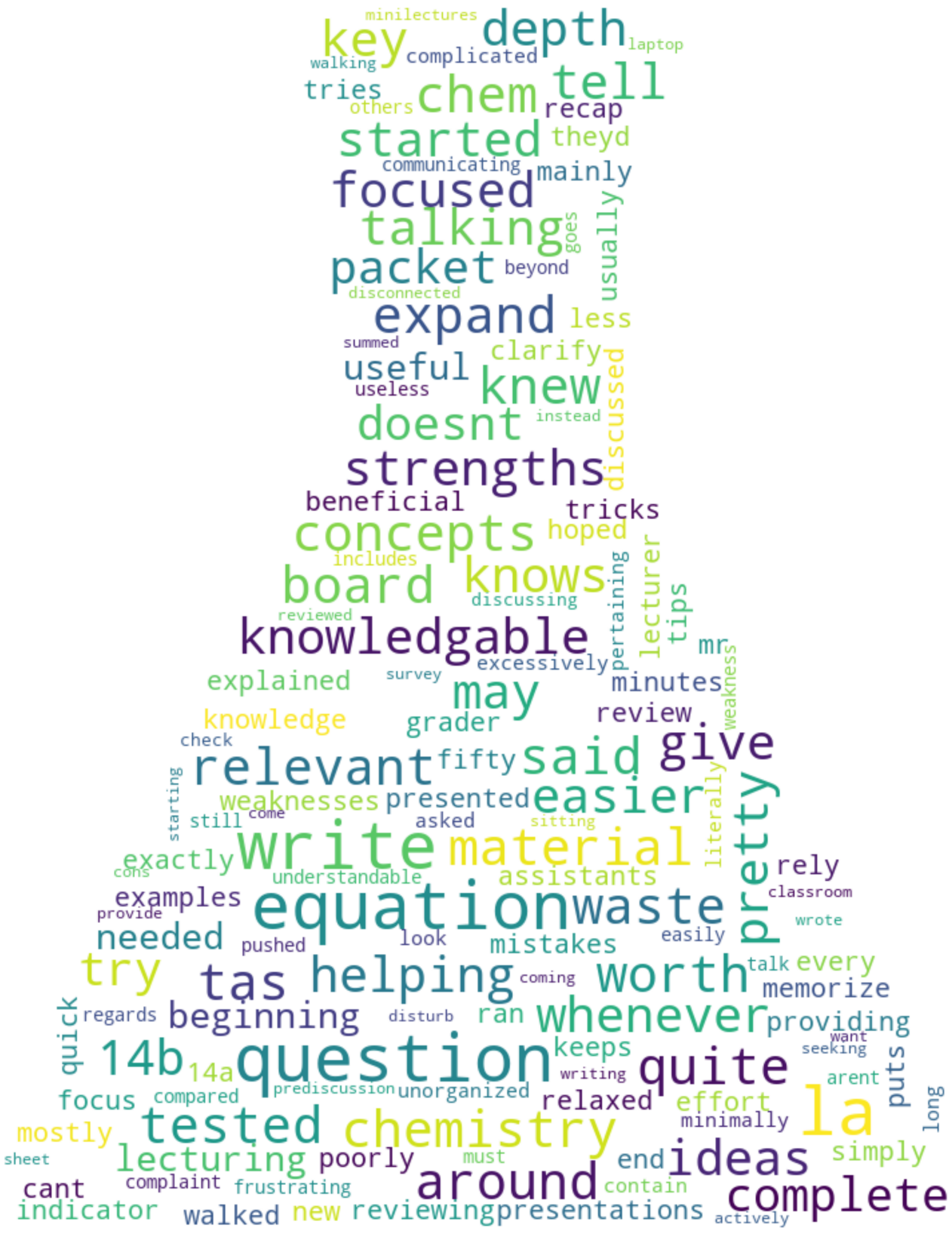

# 5 Subreport for terms associated with: [redacted]

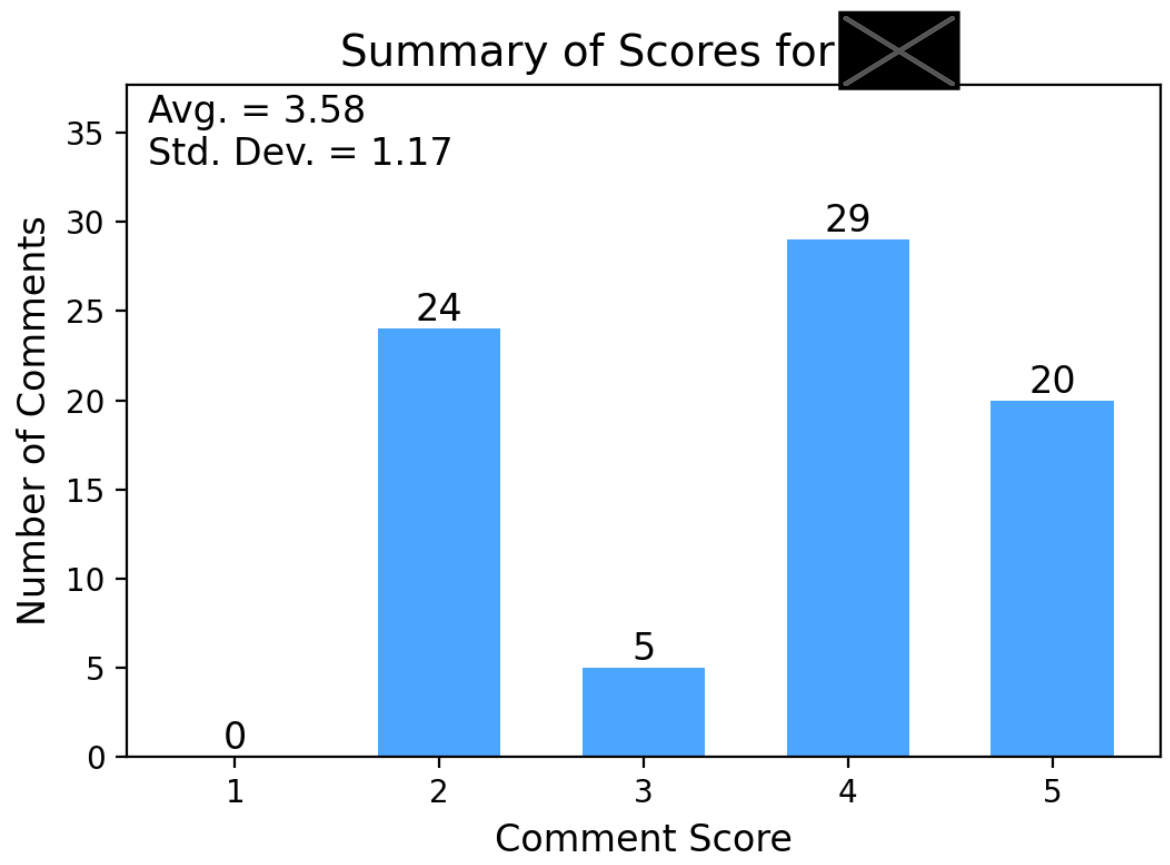

## 5.1 Table

| Review | Score |
|---|---|
| [redacted] was an outstanding ta always willing to understand our situations and expand on course material. | 5 |
| the ta was super helpful. | 5 |
| [redacted], you are by far one of my favorite ta's! | 5 |
| the ta was extremely knowledgeable in the course material, and made sure to always write on the board what from the lectures we would need to complete the discussion packet. | 5 |
| his discussion sections were very useful to me as they covered confusing topics from lecture. | 4 |
| the ta really focused on going into greater depth of what was being taught in class | 4 |
| [redacted] taught so efficiently and covered a lot of ground in a single discussion. | 4 |
| [redacted] humor also brings light to the grogginess of our 8am discussion... | 4 |
| but he did his best to try to reinforce what we learn in lecture. | 4 |
| [redacted] did his best to clarify material from the lectures. | 4 |
| the only thing i found helpful from discussion sections was what [redacted] would write on the board before class started. | 3 |
| the packet itself is the only complaint in that they were very difficult compared to the exam. | 2 |
| the course was confusing at times, so the ta helped clarify the content very well. | 2 |
| sometimes he would be a bit slow, worksheets would have helped. | 2 |
| always said they'd go over problems with us in the end | 2 |
| not too relaxed in which he was not efficient | 2 |

# 6 Subreport for terms associated with: Discussion

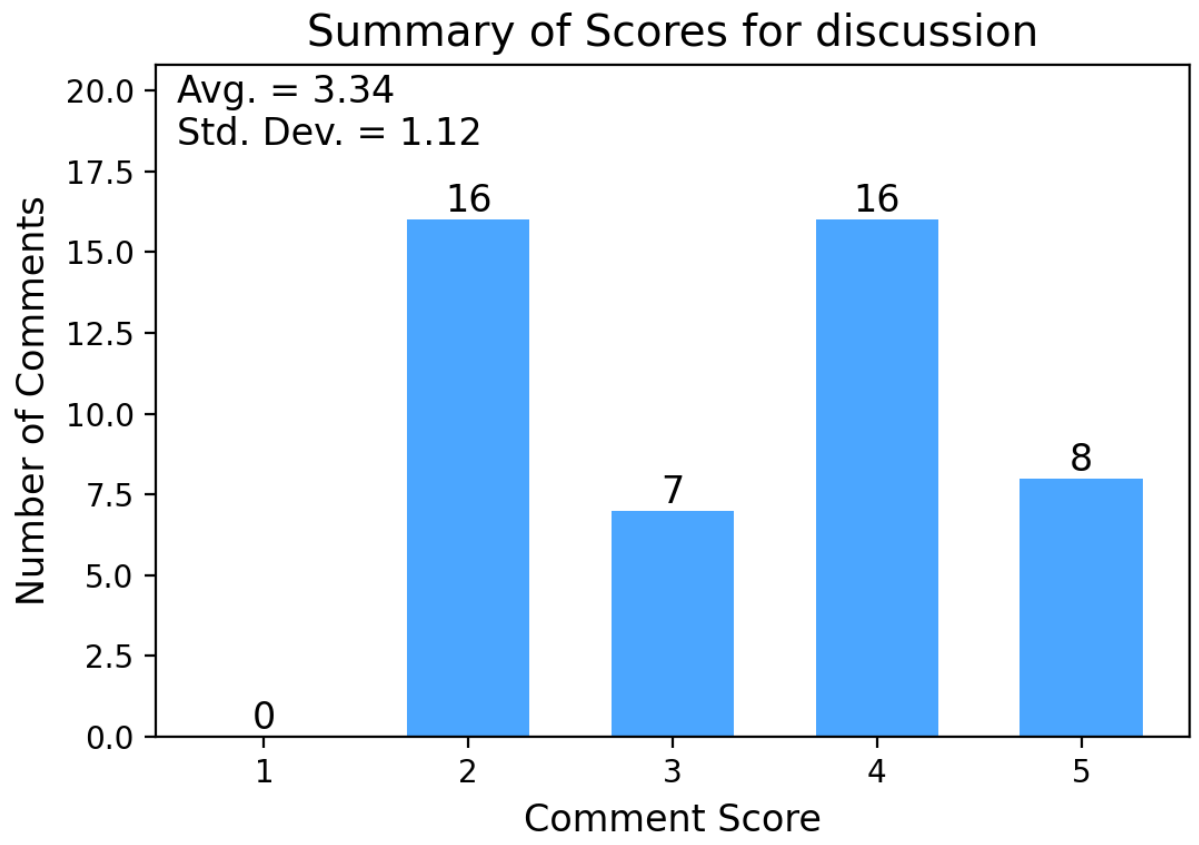

## 6.1 Table

| Review | Score |
|---|---|
| the ta was extremely knowledgeable in the course material, and made sure to always write on the board what from the lectures we would need to complete the discussion packet. | 5 |
| i felt that he genuinely cared that we knew and understood the course material, and he took the time to cover what he believed were the most challenging concepts. | 5 |
| [redacted] was an outstanding ta always willing to understand our situations and expand on course material. | 5 |
| he made sure that during discussions we were aware of what was most important especially for upcoming exams. | 4 |
| the ta was very knowledgeable about the course and advocated for student's learning via discussion sections. | 4 |
| he expanded on the course material in a way that made it much easier to understand what we were taught. | 4 |
| his discussion sections were very useful to me as they covered confusing topics from lecture. | 4 |
| the ta really focused on going into greater depth of what was being taught in class | 4 |
| you never graded unfairly like some ta's that i have met and you explained concepts to me that i found very confusing during discussion section! | 3 |
| it seemed like we only went to discussion to work on a worksheet on our own. | 3 |
| i wish we would have focused less on the discussion worksheets and more on reviewing content, since we can learn the worksheet concepts on our own once the key was posted. | 2 |
| i think he could improve on walking around the classroom and actively seeking to help students instead of students coming to him. | 2 |
| but would result in not very effective discussions as sometimes no one had real questions or just started learning a new topic. | 2 |
| the discussion worksheets should not be so difficult and should contain relevant information pertaining to exams. | 2 |
| [redacted] keeps to himself most of the time during discussions. | 2 |

# 7 Subreport for terms associated with: Homework

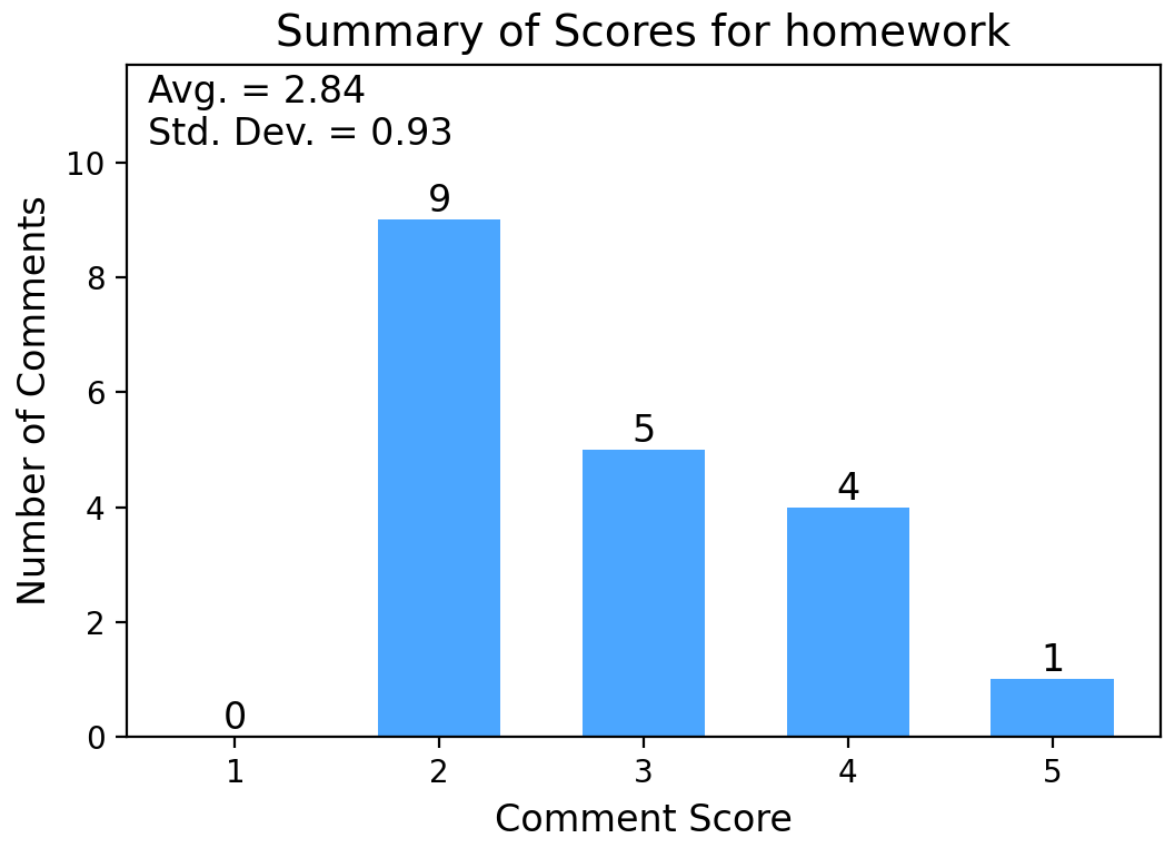

## 7.1 Table

| Review | Score |
|---|---|
| [redacted] does well in explaining concepts and ideas taught in lecture to us in more detail, providing examples and doing practice problems. | 5 |
| the discussions for this class aren't worth my time since i can easily try the worksheets on my own time and still do well in the course. | 4 |
| his pre-discussion lecture was helpful for the worksheet that day. | 4 |
| very relaxed with menial things such as turning in homework late. | 4 |
| his homework policy was highly appreciated. | 4 |
| i just think it would be more helpful to focus mainly on that rather than the worksheets, especially since i dont think the worksheets are a good indicator of what the professor tested us on. | 3 |
| i do wish he continued lecturing and explaining certain questions and concepts rather than allowing the majority of the discussion to be individual work time on the worksheets | 3 |
| the ta would expand on course ideas which was helpful i think the difficulty of the worksheets was too much | 3 |
| it seemed like we only went to discussion to work on a worksheet on our own. | 3 |
| i hoped he would go over more worksheet questions during discussion | 3 |
| the discussion worksheets were excessively long and complicated, and with a ta that never really helped, discussion sections for me were quite literally a frustrating, useless waste of my time. | 2 |
| i wish we would have focused less on the discussion worksheets and more on reviewing content, since we can learn the worksheet concepts on our own once the key was posted. | 2 |
| there wasn't enough las to help everyone so mostly discussion seemed like a waste since we could have done the worksheet on our own as homework or on our own time. | 2 |
| additionally, he was quite lazy, simply doing homework problems in class rather than explaining course material further or expanding on the professors lectures. | 2 |
| all this ta did was write equations on the board, go over said equations, and then tell us to work on the worksheet ourselves. | 2 |
| the ta could work on the interest of helping students and going through discussion worksheet problems with the class. | 2 |
| the discussion worksheets should not be so difficult and should contain relevant information pertaining to exams. | 2 |
| however, he made many mistakes during practice problems, leading to confusion. | 2 |
| sometimes he would be a bit slow, worksheets would have helped. | 2 |

# 8 Subreport for terms associated with: material

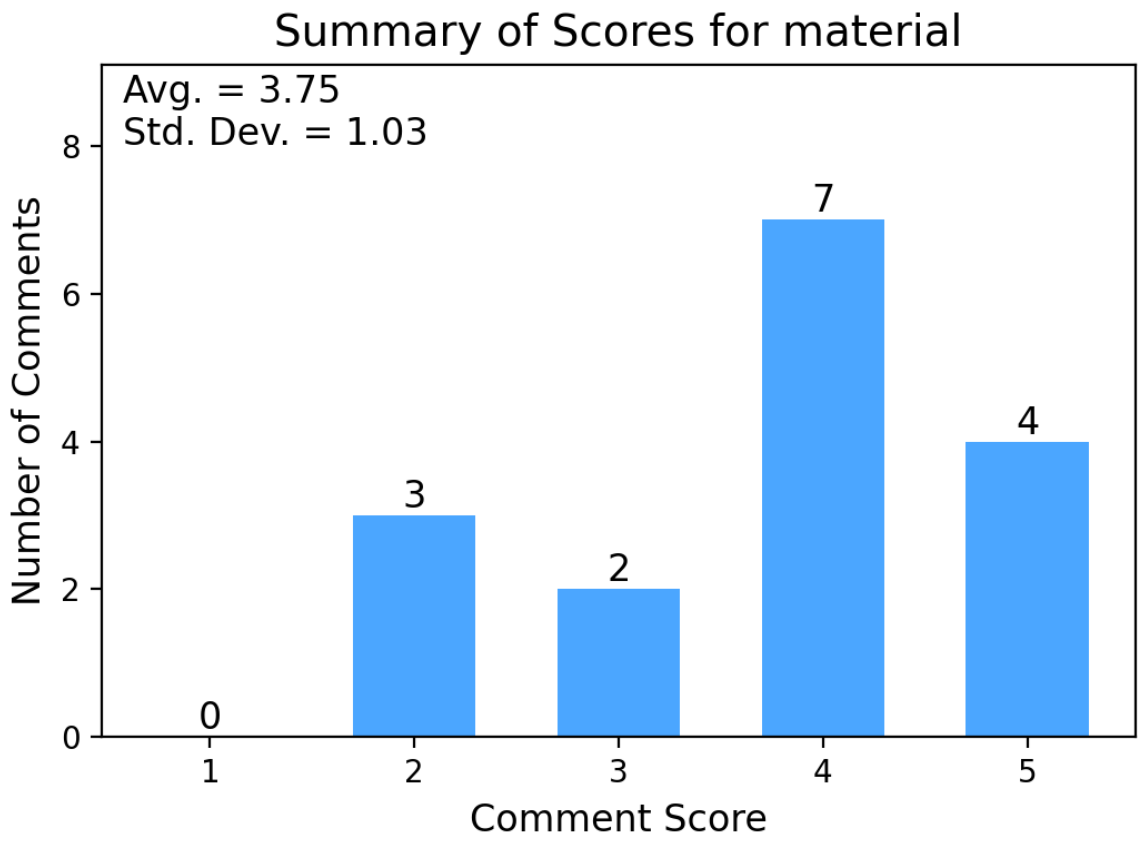

## 8.1 Table

| Review | Score |
|---|---|
| strengths: gave very useful and organized mini-lectures at the beginning of discussions that summed up what we had just learned in class and he really knew the course and its material well. | 5 |
| the ta was extremely knowledgeable in the course material, and made sure to always write on the board what from the lectures we would need to complete the discussion packet. | 5 |
| i felt that he genuinely cared that we knew and understood the course material, and he took the time to cover what he believed were the most challenging concepts. | 5 |
| [redacted] was an outstanding ta always willing to understand our situations and expand on course material. | 5 |
| i also really enjoy how he does a brief complete summary of all the notes we've learned the previous week because it sums up the new material. | 4 |
| he simplifies the complex material and teaches us exactly what we need to know, and definitely makes discussion worth attending! | 4 |
| other than that, the ta made sure to ask and help others and was very easy to talk to in regards to the course material. | 4 |
| he expanded on the course material in a way that made it much easier to understand what we were taught. | 4 |
| he was always open to taking students' questions and clearing any misconceptions about the material. | 4 |
| [redacted] did his best to clarify material from the lectures. | 4 |
| good knowledge of course material. | 4 |
| sometimes went to in depth with material not tested on/easy concepts, and other times did not go in depth enough for relevant material/more difficult concepts. | 3 |
| the presentations of material from class was mostly just writing down an equation sheet. | 3 |
| it is understandable that students must be tested and pushed to truly understand the material however, sometimes it seemed like the questions were beyond what we need to understand. | 2 |
| additionally, he was quite lazy, simply doing homework problems in class rather than explaining course material further or expanding on the professors lectures. | 2 |
| while the material presented in the lecture was difficult, the ta was able to reintroduce the material to students so that we can fully understand it. | 2 |

# 9 Subreport for terms associated with: concepts

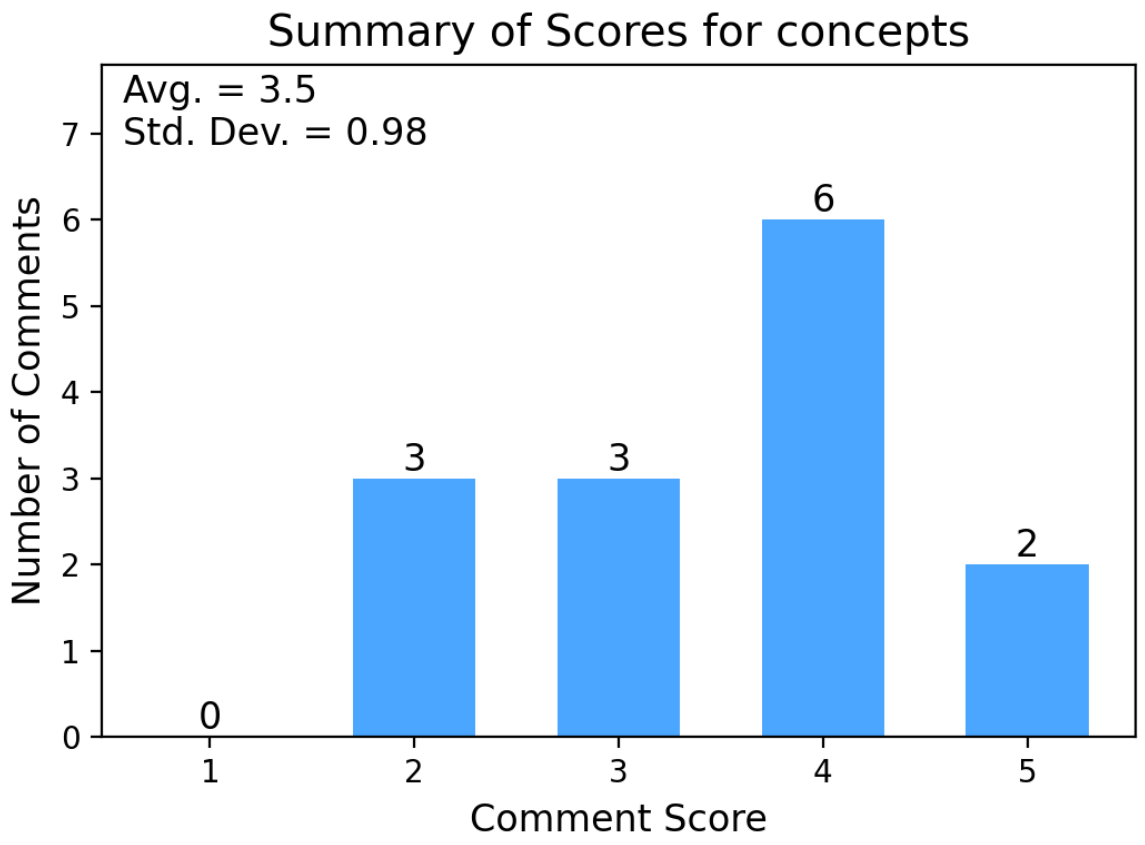

## 9.1 Table

| Review | Score |
|---|---|
| i felt that he genuinely cared that we knew and understood the course material, and he took the time to cover what he believed were the most challenging concepts. | 5 |
| [redacted] does well in explaining concepts and ideas taught in lecture to us in more detail, providing examples and doing practice problems. | 5 |
| [redacted] knows what he is talking about and provides his students with a lot of tips and tricks to memorize important concepts. | 4 |
| [redacted] is really skilled in breaking down difficult concepts and getting the most important points across! | 4 |
| he writes and go over the weekly concepts on the board during discussion which are very helpful. | 4 |
| he accommodates our level of understanding to help us work through difficult concepts. | 4 |
| [redacted] clarified many of the concepts that i was concerned about. | 4 |
| but, [redacted] did review concepts very well which helped. | 4 |
| i do wish he continued lecturing and explaining certain questions and concepts rather than allowing the majority of the discussion to be individual work time on the worksheets | 3 |
| sometimes went to in depth with material not tested on/easy concepts, and other times did not go in depth enough for relevant material/more difficult concepts. | 3 |
| you never graded unfairly like some ta's that i have met and you explained concepts to me that i found very confusing during discussion section! | 3 |
| i wish we would have focused less on the discussion worksheets and more on reviewing content, since we can learn the worksheet concepts on our own once the key was posted. | 2 |
| i think that it was hard for him to explain concepts as 14b is a very weird class with not that much interconnection | 2 |
| but i wish he went over harder concepts sometimes rather than the easier ones | 2 |

# 10 Subreport for terms associated with: questions

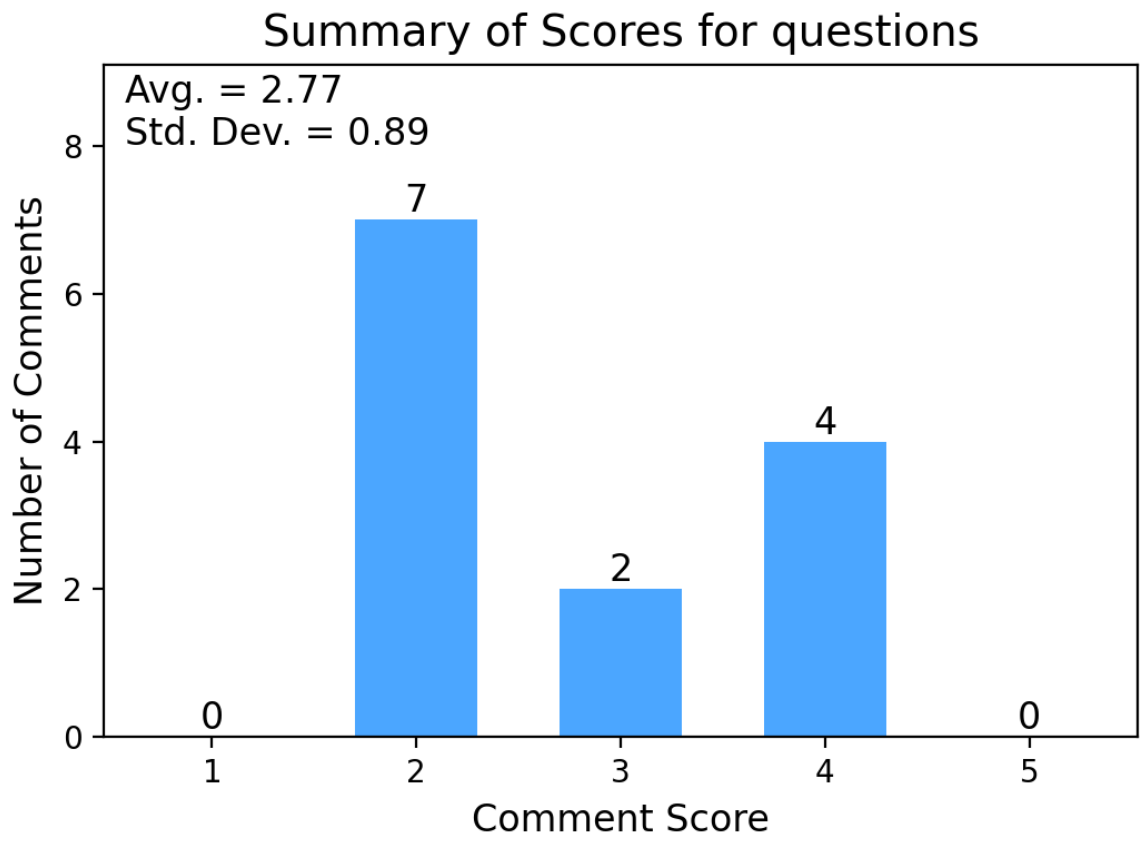

## 10.1 Table

| Review | Score |
|---|---|
| he made sure to give a quick recap of the topics discussed in lecture and walked around for questions students may have. | 4 |
| he is also extremely knowledgeable about chemistry and tries to answer as many questions as possible. | 4 |
| he was always open to taking students' questions and clearing any misconceptions about the material. | 4 |
| he is quite knowledgeable and is able to answer my questions. | 4 |
| i do wish he continued lecturing and explaining certain questions and concepts rather than allowing the majority of the discussion to be individual work time on the worksheets | 3 |
| i hoped he would go over more worksheet questions during discussion | 3 |
| i think sometimes the questions we went over were not as beneficial as some of the other times so perhaps spending more time looking for questions or asking for student suggestions may be beneficial. | 2 |
| it is understandable that students must be tested and pushed to truly understand the material however, sometimes it seemed like the questions were beyond what we need to understand. | 2 |
| but would result in not very effective discussions as sometimes no one had real questions or just started learning a new topic. | 2 |
| he doesn't really do much in discussion and doesn't expand on any topics or goes over many questions. | 2 |
| he was dry and frankly quite rude, making it difficult to ask questions or attend his office hours. | 2 |
| whenever we have questions, we usually have to rely on the learning assistants for help. | 2 |
| weakness includes not discussing many of our questions on exams. | 2 |

# 11 Subreport for terms associated with: chemistry

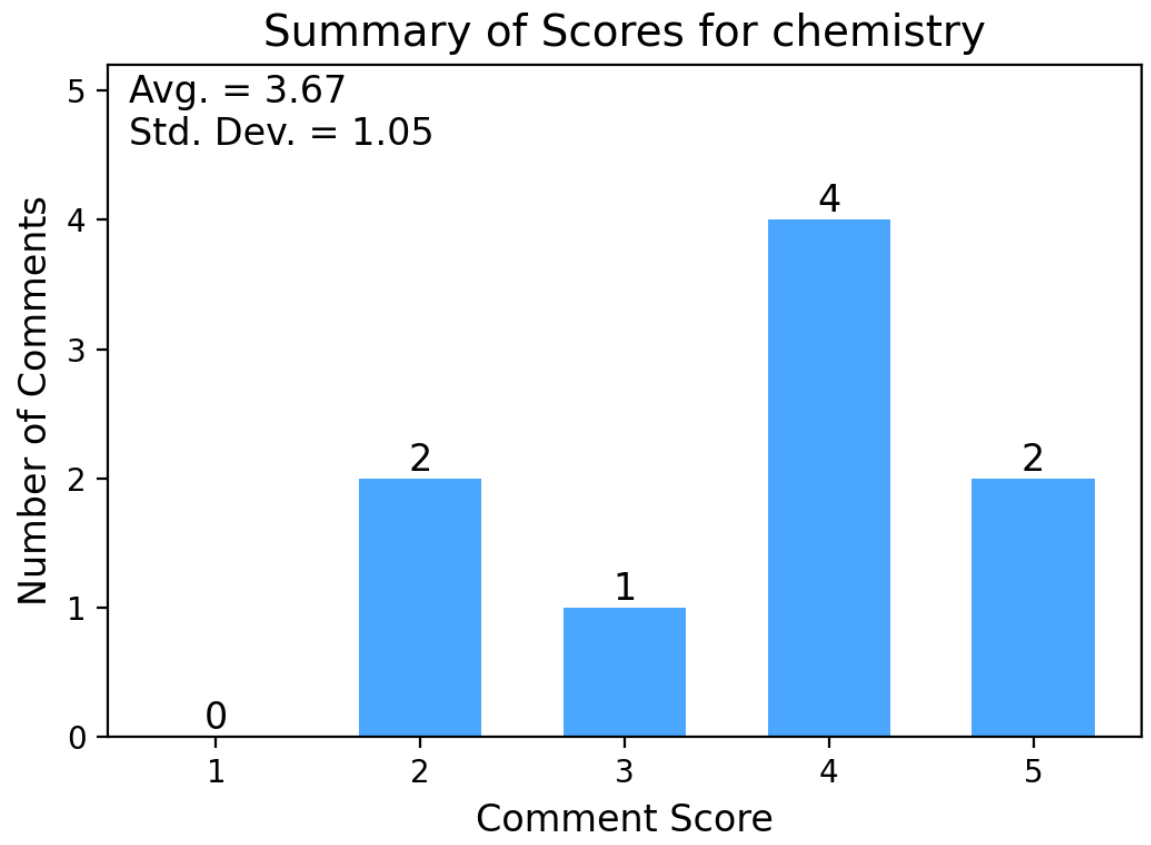

## 11.1 Table

| **Review** | **Score** |
|---|---|
| i also had him for chem 14a last quarter, and the amount of effort he puts into helping the students in his section is amazing. | 5 |
| i believe that [redacted] is honestly the best chemistry ta out there. | 5 |
| he is also extremely knowledgeable about chemistry and tries to answer as many questions as possible. | 4 |
| strengths: [redacted] really cared about the students and know a lot about chemistry. | 4 |
| he's cool and definitely knows chemistry well and explains it well | 4 |
| [redacted] is a good ta for chem 14b. | 4 |
| it is difficult to teach a week's worth of chemistry in fifty minutes | 3 |
| my teaching assistant for chem 14b, [redacted], was extremely unapproachable and unfriendly as a ta. | 2 |
| i wish i went to his office hours more often in chem 14a | 2 |

# 12 Subreport for terms associated with: Exam

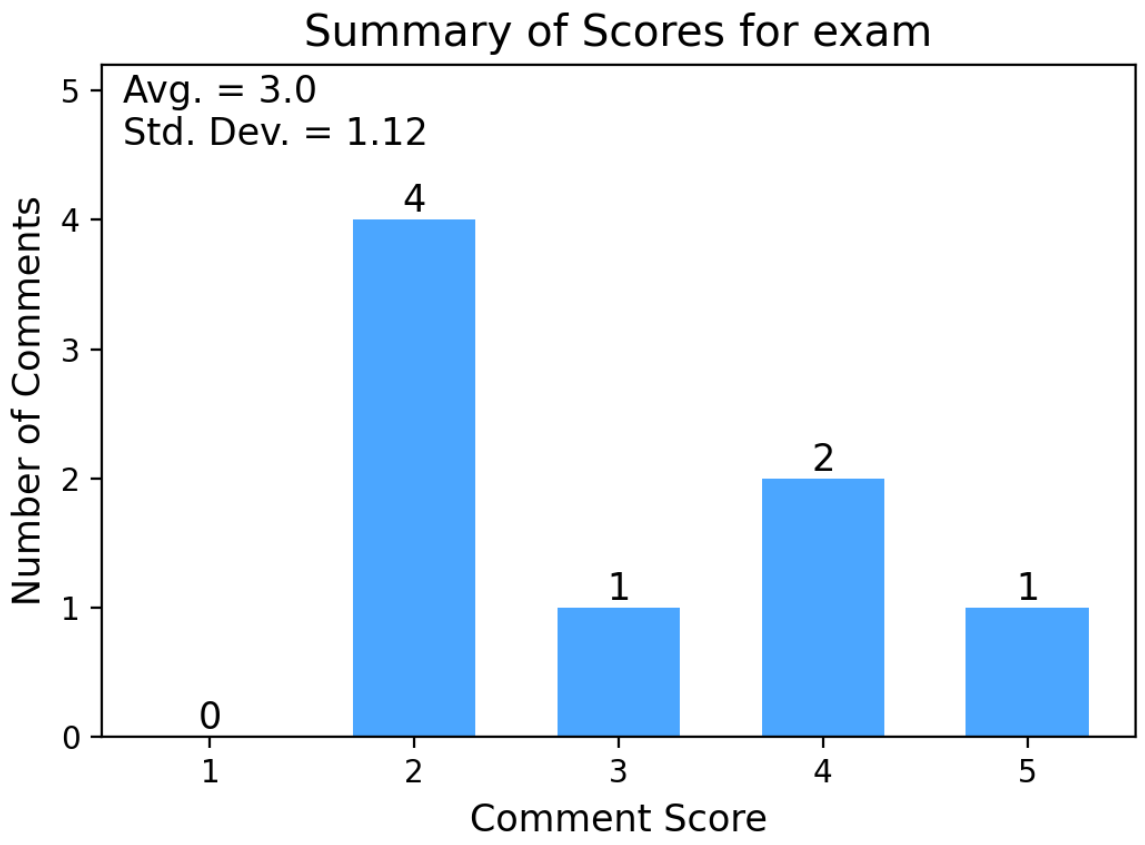

## 12.1 Table

| Review | Score |
|---|---|
| [redacted] does well in explaining concepts and ideas taught in lecture to us in more detail, providing examples and doing practice problems. | 5 |
| he made sure that during discussions we were aware of what was most important especially for upcoming exams. | 4 |
| very good overview of topics presented in lectures and relevant examples. | 4 |
| he acknowledges what will be on the exam and how difficult each subject is. | 3 |
| the discussion worksheets should not be so difficult and should contain relevant information pertaining to exams. | 2 |
| the packet itself is the only complaint in that they were very difficult compared to the exam. | 2 |
| did not waste time going over things that we already knew or would not be on exams. | 2 |
| weakness includes not discussing many of our questions on exams. | 2 |

# 13 Subreport for terms associated with: board

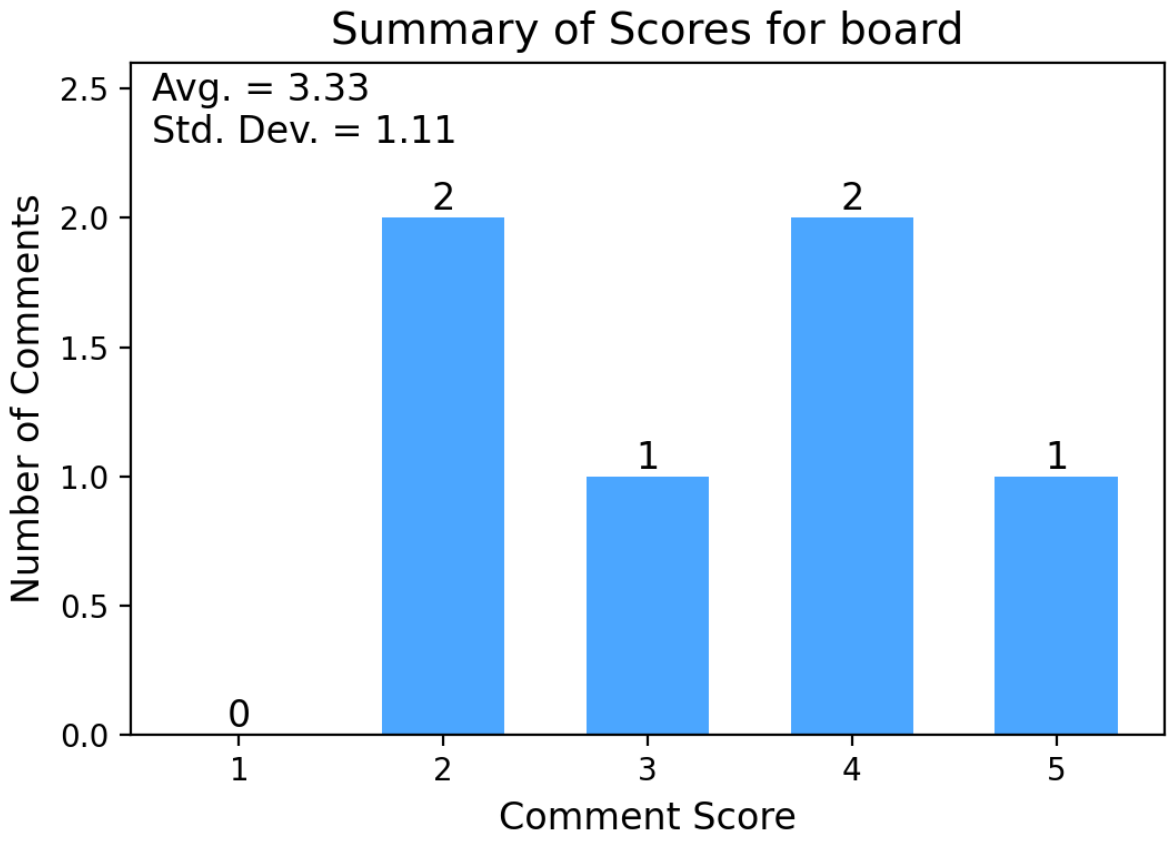

## 13.1 Table

| Review | Score |
|---|---|
| the ta was extremely knowledgeable in the course material, and made sure to always write on the board what from the lectures we would need to complete the discussion packet. | 5 |
| i really enjoyed how he reviewed key topics before starting the survey and wrote important equations on the board. | 4 |
| he writes and go over the weekly concepts on the board during discussion which are very helpful. | 4 |
| the only thing i found helpful from discussion sections was what [redacted] would write on the board before class started. | 3 |
| i stopped going because he was simply just doing problems from the book on the board for fifty minutes, a huge waste of my time. | 2 |
| all this ta did was write equations on the board, go over said equations, and then tell us to work on the worksheet ourselves. | 2 |

# 14 Subreport for terms associated with: knew

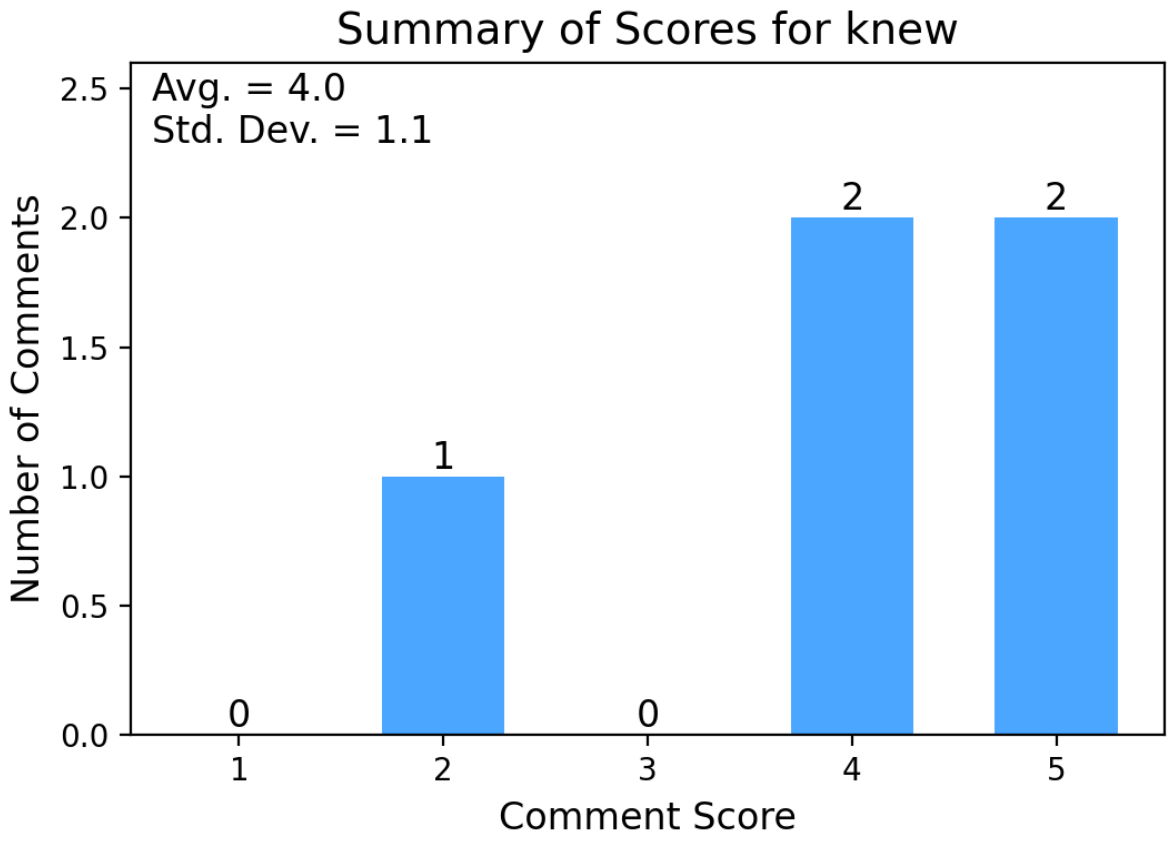

## 14.1 Table

| Review | Score |
|---|---|
| strengths: gave very useful and organized mini-lectures at the beginning of discussions that summed up what we had just learned in class and he really knew the course and its material well. | 5 |
| i felt that he genuinely cared that we knew and understood the course material, and he took the time to cover what he believed were the most challenging concepts. | 5 |
| knew exactly what we needed help with at any time. | 4 |
| he really knew what he was talking about. | 4 |
| did not waste time going over things that we already knew or would not be on exams. | 2 |

# 15 Subreport for terms associated with: Office hour

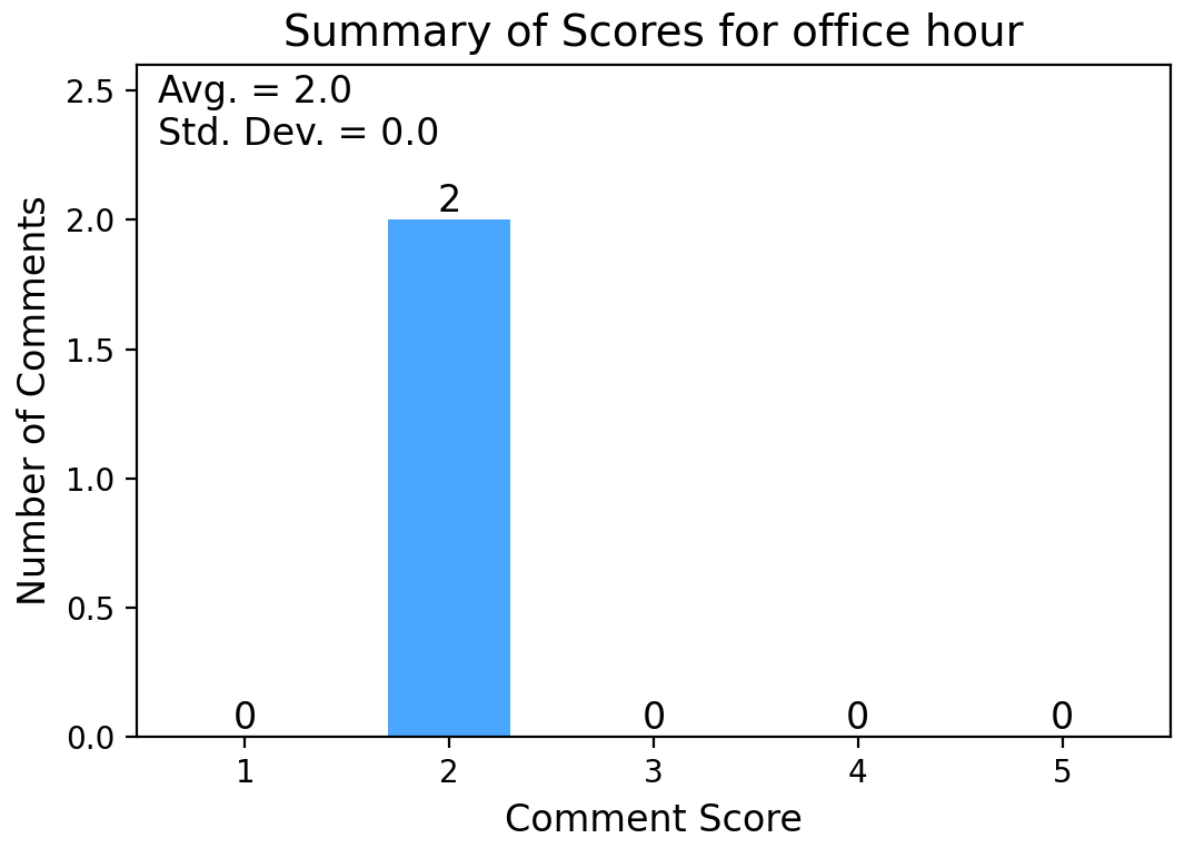

## 15.1 Table

| **Review** | **Score** |
|---|---|
| he was dry and frankly quite rude, making it difficult to ask questions or attend his office hours. | 2 |
| i wish i went to his office hours more often in chem 14a | 2 |

# 16 Subreport for terms associated with: Book

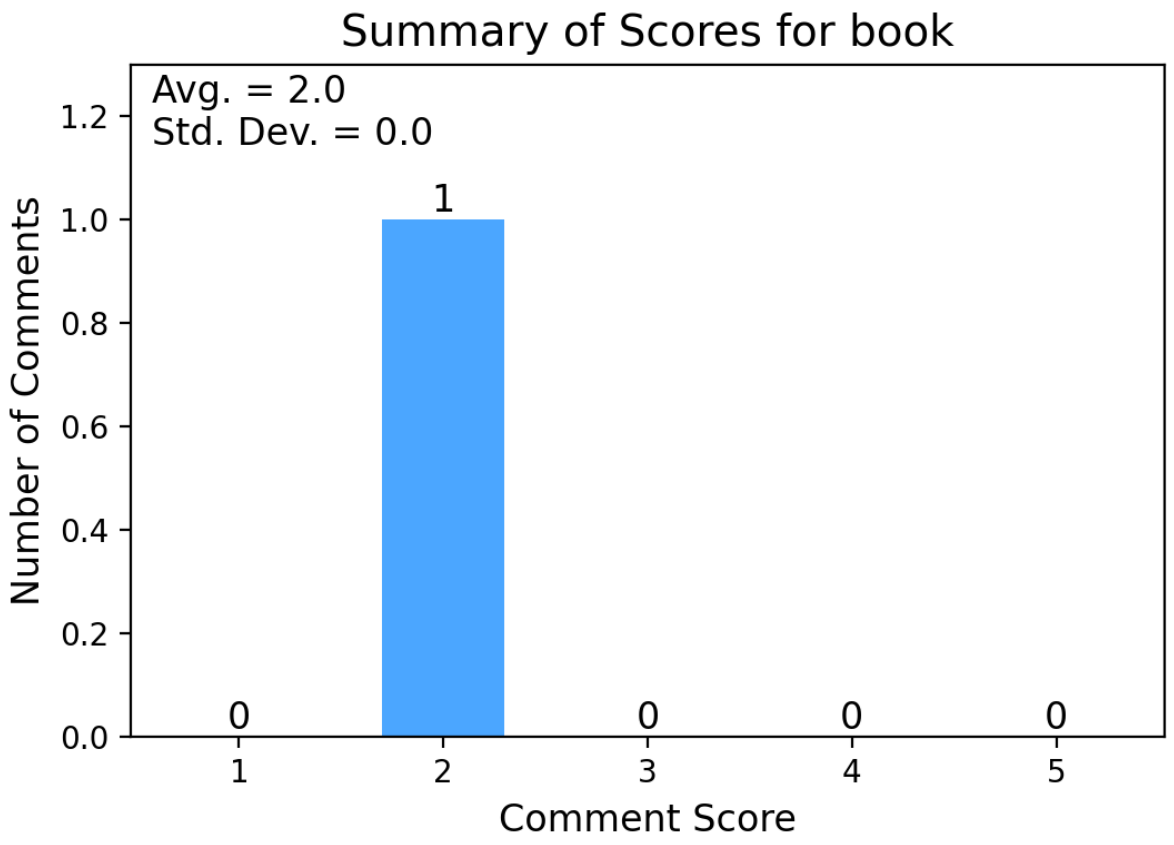

## 16.1 Table

| **Review** | **Score** |
|---|---|
| i stopped going because he was simply just doing problems from the book on the board for fifty minutes, a huge waste of my time. | 2 |